# SYNTAGMA. The Lexical Database

Daniel Christen

**Abstract**

This paper discusses the structure of Syntagma's Lexical Database (focused on Italian). The basic database consists in four tables. Table Forms contains word inflections, used by the POS-tagger for the identification of input-words. Forms is related to Lemma. Table Lemma stores all kinds of grammatical features of words, word-level semantic data and restrictions. In the table Meanings meaning-related data are stored: definition, examples, domain, and semantic information. Table Valency contains the argument structure of each meaning, with syntactic and semantic features for each argument. The extended version of SLD contains the links to Syntagma's Semantic Net and to the WordNet synsets of other languages.

The main resource of the Syntagma parser[1] is its lexical database: the Syntagma Lexical Database (SLD). The current SLD addresses only Italian, but is related to Word Net's *synsets*. SLD has been generated through automatic procedures by a program, called Dictionary Scanner, which extracted and classified all kinds of lexical information coming from the source dictionary (given in htm format).

## 1. SLD structure

The SLD consists of four tables, related as follows:

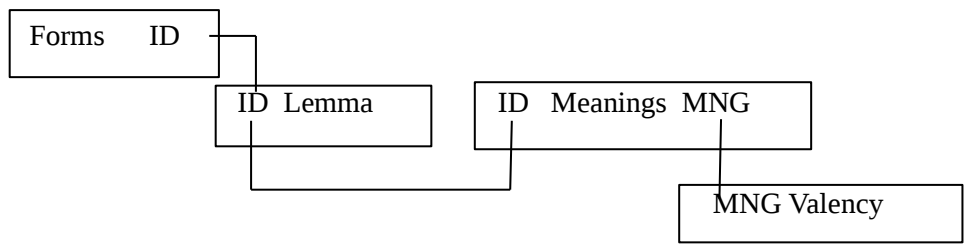

### 1.1. Forms
The table *Forms* contains the inflected forms of words, including altered forms. Each form is provided with its morphological data (verbal mood, tense, person, genre, number), and is related to a lexical item (a record) of the *Lemma* table (1.2 below), identified by the ID index.
*Forms* is used by the POS-tagger for the identification of input-words.

### 1.2 Lemma
The table *Lemma* shows the basic form of a word, with its phonetical and grammatical proprieties (category, conjugation, gender and number). Also word-level features, such as etymology and aliases, are stored here.
There may be more than one lemma with the same basic-form and the same ID, but belonging to a different category: for example nouns that can also be used as adjectives or vice versa ("abbagliante").
Each lemma has one or more meanings which become its sense in a given context.

---

[1] D. Christen, *Syntagma. A Linguistic Approach to Parsing.* (2013), http://arxiv.org/abs/1303.5960.

**1.3 Meanings**

Meanings are related to Lemmas through the ID index. Each entry, identified by the MNG index, contains the information that belongs to one specific meaning of a word, given by its dictionary definition. Beside the definition, all kinds of meaning-related information are stored here. First of all, grammatical proprieties, like transitivity, reflexivity, auxiliary and control; but also morphological restrictions over mood, tense, person, genre, number that come with a specific meaning.

The remaining fields are dedicated to data which are used in Word Sense Disambiguation tasks and during the selection of parsing results, where the Syntagma parser applies a close interaction between syntactic and semantic information[2].

These features inherit the use of a word, its register, its domain ("biology", "economy", "medicine"...); but also its synonyms and contraries (if mentioned in the source-dictionary) and semantic features ("modal", "espistemic", "causative", "perceptive"...).

The content of a Meaning data addresses also locutions and idiomatic expressions in which the given word appears.

Examples and quotations, coming from the source-dictionary and which illustrate the use of a given word-meaning in its context, are also stored here.

The following example shows the record *Meaning* 23293.01, which contains information about the first meaning of the verb "forget" (Ital.: "dimenticare"):

> MNG 23293.01; LEX "dimenticare"; PRON "[di-men-ti-cà-re]"; ID 23293;
> CAT 100; CATS "v.tr."; AUX "avere"; TRN 2; RFL 0; CTRL F; POS ; LNG; REG ;
> ALIAS ;
> SYN ("scordare" "obliare"); CNTR ; IDIOM ; SEM ; REF ; RESTR ; VL "(subj-v-arg)";
> PRF "Perdere la memoria di una cosa";

The record *Meaning* also contains a field with syntactic constraints or other kinds of restrictions, related to a specific word meaning. Constraints have the form of a kind of logic expression, resulting from a formalization, done automatically by the Dictionary Scanner, which thus translates additional information given in the source-dictionary. Constraints allow AND and OR operators and emebedding.
**For example,** meaning 995.67 of the word "acqua" ("water") means in Italian "amniotic fluid" ("liquido organico, liquido amniotico". The information that, in this specific sense, the word "acqua" can appear only in the plural form ("in partic., al pl.,"), has been transposed by the Dictionary Scanner in the following constraint:

> RESTR "{Context(TAG=NP) Target(TAG=N) Restriction(NUM=pl.)}"

The *Meaninig*-record of a word provided with an argument structure is related to the *Valency* table through the index MNG.

**1.4 Valency**

The *Valency* table describes the argument structure of a specific word meaning (*Meaning* record). For each argument, its syntactic function (FNCT), its projection (NP, VP, AdjP, AdvP, Clause), the required connectives, its optionality and different kinds of constraints that may occur are stored. Constraints may be over lexical (VLEX) or semantic (VSEM) features of arguments: for example a specific sense of "abbaiare" ("bark") shows that its subject has to be a "dog", while other meanings of the same verb, i.e. figuratively, also allow a "person" as a subject. Constraints may also specify the syntactic context and word position on the expression level.

Varieties of the same function are also registered, for example when an object may have both the projection as an NP or as a dependent Clause (i.e. verb "see").

For example, MNG 23293.01 the verb "dimenticare" ("forget") belongs, in the dictionary, to a basic

---
[2]  D. Christen, *Syntagma. A Linguistic Approach to Parsing*, 2.2.6, p.5, and 2.8.2, pp. 10-13.

subj-verb-obj structure. But it has the following enhanced argument structure in the table *Valency*:

    [1;1; FNCT subj; VCAT NP; VLEX; RGG; OPT F; VMDV; VTMP; VPRS; VGEN; VNUM; VPOS; VSEM; VREF; VRESTR;]

    [2;1; FNCT v; VCAT VP; VLEX; RGG; OPT F; VMDV; VTMP; VPRS; VGEN; VNUM; VPOS; VSEM; VREF; VRESTR;]

    [3;1; FNCT obj; VCAT NP; VLEX; RGG; OPT F; VMDV; VTMP; VPRS; VGEN; VNUM; VPOS; VSEM; VREF; VRESTR;]

    [3;2; FNCT obj; VCAT C; VLEX; RGG ("di"); OPT F; VMDV (inf); VTMP; VPRS; VGEN; VNUM; VPOS; VSEM; VREF; VRESTR;]

    [3;3; FNCT obj; VCAT C; VLEX; RGG ("che"); OPT F; VMDV (ind cnd cong); VTMP; VPRS; VGEN; VNUM; VPOS; VSEM; VREF; VRESTR;]

    [3;4; FNCT obj; VCAT C; VLEX; RGG ("di"); OPT F; VMDV; VTMP; VPRS; VGEN; VNUM; VPOS; VSEM; VREF; VRESTR;]

This states that the syntactic object function (FNCT=obj) may have four types of projection: as an NP; as an infinite clause (VCAT=C) (VMDV=inf), introduced by a preposition (RGG="di"); a finite clause (MDV=indicative or conditional or subjuntive), introduced by the connective (RGG="che"); or a PP -which Syntagma treats as an NP- introduced by the preposition "di".
The two C projections of the object function are the result of the formalization, done by the Dictionary Scanner, of the information that this function may be expressed also in the form of a clause with the given features ("*anche con l'arg. espresso da frase (introd. da di o da che)*".

If the source-dictionary mentions some additional or optional argument, the related expression found in the dictionary is formalized and added to the argument structure.

**2. SLD dimensions** (language : Italian)

Forms:         1'500'000  (aproxim.)
Lemmas:        90'000  (aproxim)
Meanings:     120'000  (aproxim)
Closed Classes:    1'172

**3. Internal links**

Some words in the table *Lemmas* are linked to other records. Through these links they are connected to the meanings of the related word.
For example, the form (table *Forms*) "abbadia" relates to the lemma (ID=37) "abbadia", that is linked to another lemma, ID=7840 "badia":

    Forms:
        37; f.; sing.; "abbadia"; "lnk";
        37; f.; pl. ; "abbadie"; "lnk";

    Lemmas:
        37; 11; "abbadia"; "s.f."; f.; sing.; "badia";

        7840; 3799; "badia"; "[ba-dì-a]"; "s.f."; f.; sing.;

## 4. Link to Syntagma's Semantic Net

The records of the *Meaning* table are linked to the entries of Syntagma's semantic net[3].

## 5. Link to Word Net

The records of the *Meaning* table are linked to the synsets of the semantic net WordNet. These links have been established through WSD procedures that will be discussed in a forthcoming paper.

## 6. Data examples

### 6.1 Forms

> 5238; 1; 1; 1; 0; 1; "bramo"; "";
> 5238; 1; 1; 2; 0; 1; "brami"; "";
> 5238; 1; 1; 3; 0; 1; "brama"; "";
> 5238; 1; 1; 1; 0; 2; "bramiamo"; "";
> 5238; 1; 1; 2; 0; 2; "bramate"; "";
> 5238; 1; 1; 3; 0; 2; "bramano"; "";
> 5238; 1; 2; 1; 0; 1; "bramavo"; "";

### 6.2 Meanings and Valency

The example shows three different meanings of the verb "abbagliare" ("dazzle"/"bedazzle"): in its transitive use, meant literally and figuratively, and in its intransitive use:

MNG 40.01; LEX "abbagliare"; LEX2; PRON "[ab-ba-glià-re]"; DISC 40; WN; CAT 100; CATS "v.tr.";
    GEN 0; NUM 0; MORF; AUX "avere"; TRN 2; RFL 0; CTRL F; REG; ALIAS; SYN; CNTR;
    IDIOM; SEM; REF; RESTR; VL "[subj-v-arg]";
    PRF "Colpire qlcu. con una luce intensa, offuscandogli momentaneamente la capacità visiva";
    EXE ("abbagliare un automobilista" "la lampada mi abbaglia"); ETM; CIT; MLG;

    [1;1; FNCT subj; VCAT NP; VLEX; RGG; OPT F; VMDV; VTMP; VPRS; VGEN; VNUM; VPOS;
    VSEM; VREF; VRESTR;]
    [2;1; FNCT v; VCAT VP; VLEX; RGG; OPT F; VMDV; VTMP; VPRS; VGEN; VNUM; VPOS;
    VSEM; VREF; VRESTR;]
    [3;1; FNCT arg; VCAT NP; VLEX; RGG; OPT F; VMDV; VTMP; VPRS; VGEN; VNUM; VPOS;
    VSEM ("PERSON"); VREF; VRESTR;]

MNG 40.02; LEX "abbagliare"; LEX2; PRON "[ab-ba-glià-re]"; DISC 40; WN; CAT 100; CATS "v.tr.";
    GEN 0; NUM 0; MORF; AUX "avere"; TRN 2; RFL
    0; CTRL F; REG "fig."; ALIAS; SYN; CNTR; IDIOM; SEM "FIG"; VL "[subj-v-arg]";
    PRF "Affascinare qlcu. provocando ammirazione"; DOM; REL; LNK;
    EXE "l'arte rinascimentale ha abbagliato i posteri"; ETM; CIT; MLG;

    [1;1; FNCT subj; VCAT NP; VLEX; RGG; OPT F; VMDV; VTMP; VPRS; VGEN; VNUM;
    VPOS; VSEM; VREF; VRESTR;]

---

3    D. Christen, *Rule-Based Semantic Tagging. An Application Undergoing Dictionary Glosses.* (2013). http://arxiv.org/abs/1305.3882. The Syntagma semantic net is available at http://www.lector.ch.

[2;1; FNCT v; VCAT VP; VLEX; RGG; OPT F; VMDV; VTMP; VPRS; VGEN; VNUM; VPOS; VSEM; VREF; VRESTR;]
    [3;1; FNCT arg; VCAT NP; VLEX; RGG; OPT F; VMDV; VTMP; VPRS; VGEN; VNUM; VPOS; VSEM ("PERSON"); VREF; VRESTR;]

MNG 40.04; LEX "abbagliare"; LEX2; PRON "[ab-ba-glià-re]"; DISC 40; WN; CAT 100; CATS "v.intr."; GEN 0; NUM 0; MORF; AUX "avere"; TRN 1; RFL 0; CTRL F; REG; ALIAS; SYN; CNTR; IDIOM; SEM; REF; RESTR; VL "[subj-v]"; PRF "Emettere una luce intensa, tanto da confondere la vista"; DOM; REL; LNK; EXE "il sole di mezzogiorno abbaglia"; ETM; CIT; MLG;

    [1;1; FNCT subj; VCAT NP; VLEX; RGG; OPT F; VMDV; VTMP; VPRS; VGEN; VNUM; VPOS; VSEM; VREF; VRESTR;]
    [2;1; FNCT v; VCAT VP; VLEX; RGG; OPT F; VMDV; VTMP; VPRS; VGEN; VNUM; VPOS; VSEM; VREF; VRESTR;]

All this information has been extracted, classified and formalized starting from the following dictionary entry:

abbagliare
[ab-ba-glià-re] v. (*abbàglio* ecc.)
- • v.tr. [sogg-v-arg]
- 1 Colpire qlcu. con una luce viva, offuscandogli la vista: *a. un automobilista*
- 2 fig. Affascinare qlcu.: *l'arte rinascimentale ha abbagliato i posteri*; ingannare, abbacinare qlcu.: *la sua eloquenza abbaglia molte persone*
- • v.intr. (aus. *avere*) [sogg-v] Emettere una luce intensa, che confonde la vista